# Towards SAR Automatic Target Recognition: Multi-Category SAR Image Classification Based on Light Weight Vision Transformer

Guibin Zhao
*College of Engineering and Physical Sciences*
*Khalifa University*
Abu Dhabi, United Arab Emirates
100060512@ku.ac.ae

Pengfei Li
*School of Information and Communication Engineering*
*University of Electronic Science and Technology of China*
Chengdu, China
lipengfei__uestc@163.com

Zhibo Zhang
*School of Systems and Computing*
*University of New South Wales*
Canberra, Australia
zhibo.zhang3@unsw.edu.au

Fusen Guo
*School of Science, Computing, and Engineering Technologies*
*Swinburne University of Technology*
Melbourne, Australia
dobbysen430@gmail.com

Xueting Huang
*Faculty of Engineering*
*University of Canterbury*
Christchurch, New Zealand
milasnow0326@gmail.com

Wei Xu
*Independent Researcher*
Los Altos, California, USA
williamxw09@gmail.com

Jinyin Wang
*Independent Researcher*
Jersey City, New Jersey, USA
jinyinsbu@gmail.com

Jianlong Chen
*Independent Researcher*
Beijing, China
jianlong.chen@ieee.org

*Abstract*—Synthetic Aperture Radar has been extensively used in numerous fields and can gather a wealth of information about the area of interest. This large-scene data-intensive technology puts a high value on automatic target recognition (ATR) which can free the utilizers and boost the efficiency. Recent advances in artificial intelligence have made it possible to create a deep learning-based SAR ATR that can automatically identify target features from massive input data. In the last 6 years, intensive research has been conducted in this area, however, most papers in the current SAR ATR field used recurrent neural network (RNN) and convolutional neural network (CNN)-varied models to deepen the regime's understanding of the SAR images. To equip SAR ATR with updated deep learning technology, this paper tries to apply a lightweight vision transformer (LViT)-based model to classify SAR images. The entire structure was verified by an open-accessed SAR data set and recognition results show that the final classification outcomes are robust and more accurate in comparison with referred traditional network structures without even using any convolutional layers.

## I. INTRODUCTION

Synthetic Aperture Radar (SAR), a prominent modern microwave sensor technology, has made substantial contributions to both civilian and military fields because of its capability to image the region where the interested targets conceal themselves. The SAR sensor system can operate in most situations independent of changes in lighting conditions or weather which can greatly impact conventional sensor regimes like infrared and optical systems. In terms of depicting targets, the fact that SAR gathers and analyzes electromagnetic data rather than employing a direct image method also distinguishes it from other common sensor systems. Due to all these special characteristics, SAR is capable of containing more compact information about the interested targets and is widely applied in modern imagery.

To increase the efficacy and flexibility of SAR ATR while reducing its complexity, the deep learning-based SAR ATR has been introduced which completely employs the power of the computer in discovering the intrinsic relationship between the input data and expected output via optimizing network parameters. The amount of human power required by this kind of SAR ATR approach is much reduced, and it is better equipped to handle input alternation like size reshape and rotation. With the advent of deep learning-based SAR ATR, the work has gradually moved from creating complex feature-extraction methods to constructing powerful network structures, and the effectiveness of those structures can be evaluated via the Moving and Stationary Target Acquisition and Recognition (MSTAR) program [1]. Throughout these years, many network structures have been conducted, but most of them focused on proposing structures based on traditional CNN or RNN to deepen the network's understanding- ing towards the MSTAR data set. For example, S. Deng et al. applied an enhanced autoencoder CNN to recognize the data set and achieved a better performance than the conventional autoencoder [2]. Z. Huang et al. completed the multi-categorial recognition task using an enhanced CNN with a designed feedback bypass [3]. In 2018, Pei, J. et al. proposed a CNN-based expandable 'multi-view' structure [4] which was further modified to an RNN-based structure in 2021 [5]. In 2019, Z. Zhou et al. used a multi-level reconstruction methodology [6]. Later in 2022, X. Ma et al. proposed a generative adversarial network (GAN) based structure [7]. J. Ai et al. proposed a multi-kernel size feature fusion CNN (MKSFF-CNN) [8].

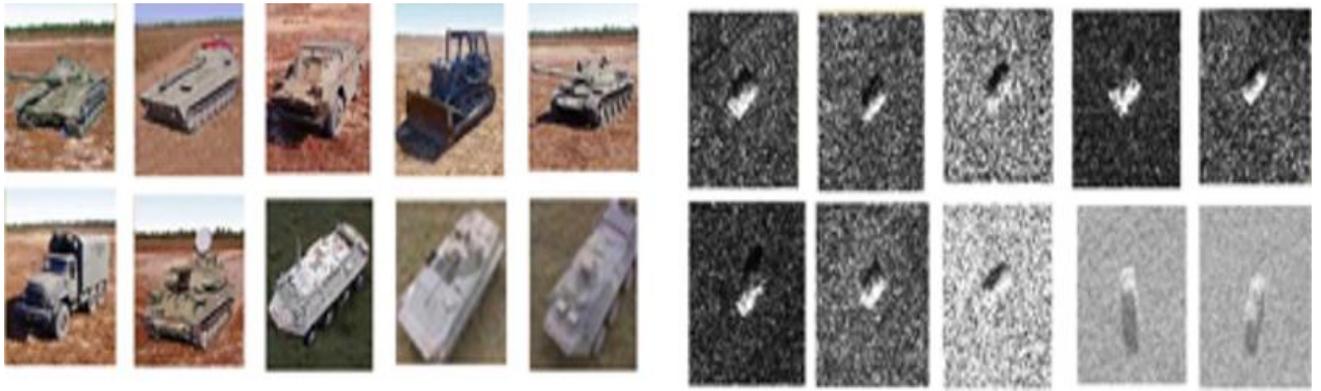

Fig. 1. The ten categories of targets in the MSTAR data set with a one-to-one Optical-SAR image match.

To introduce the updated knowledge of the deep learning field into SAR ATR, this paper plans to classify the MSTAR via a vision transformer-based structure [9] [10] [11] which is encouraged by the multi-head self-attention mechanisms [12]. It is also noticeable that we deduced the size of the original ViT to make it more compatible with the current data set scale and formed a lightweight ViT (LViT), and this structure can be expanded along with the data size to classify other SAR sets especially those internally-collected ones with more images.

To integrate the latest advancements in deep learning into the area of Synthetic Aperture Radar (SAR) Automatic Target Recognition (ATR), the main contributions of this research paper are mentioned as follows:

- This paper proposes a novel approach or methodology, possibly involving a lightweight vision transformer (LViT)-based model for Synthetic Aperture Radar (SAR) image classification.

- This paper presents a comparison of this model's performance against traditional network structures, demonstrating improved accuracy and robustness in automatic target recognition in SAR data.

- The paper introduces a new framework for processing SAR images, which could be an advancement in the field of remote sensing.

- The findings of this paper provide valuable experience in terms of practical applications or implications of this research in relevant fields, such as military, aerospace, or environmental monitoring.

The rest of this paper is structured as follows: Section 2 introduces the utilized MSTAR data set and the proposed method. The results and further analysis of the proposed framework are discussed in Section 3. This paper is concluded in Section 4.

## II. DATA SET AND PROPOSED METHODOLOGY

### A. MSTAR Data Set

The MSTAR project includes ten types of targets, such as 2 varieties of tanks, 4 categories of armored vehicles, trucks, anti-aircraft units, bulldozers, and howitzers. Figure 1 provides more information about these targets through the paired SAR and its optical images. In the MSTAR project, because all information was gathered by an aircraft with SAR imaging capability that can scan the region of interests (ROI) from above, it can be discovered that great similarities are observed among different target classes even though the ground imageries of those targets are in relatively high quality. This method of data acquisition offers a unique perspective and challenges for image classification and recognition. Despite the high quality of ground imagery, there are significant similarities observed among different target classes. This similarity poses a challenge in differentiating between the various categories. In spite of this, there is also a depression angle variation between the train (17°) and test (15°) data sets, this alternation also puts a higher requirement on the generalization ability of the constructed network since the train-and-test sets are not split from a single data set under the same circumstance.

### B. Model Architecture

The main body of the model architecture follows the structure of the original Vit model introduced by [9] [10]. This LViT and MSTAR-combined problem can also be viewed as a meta-learning task with a K-way and n-shot set, where K ranges from 1 to 10 classes and n ranges from 155 to 573 data points (depends on the inclusion of MSTAR categories and the split of query and support sets). The overview of architecture is depicted in Fig.2. The model is composed of two components: a transformer encoder and MLP. We first process the images with a vision transformer encoder and then feed the output features into MLP for the ten-category classification task.

For the input image as $\exists x \in \mathbb{R}^{48 \times 48 \times 1}$, 1 denotes one channel and (48,48) denotes the width and height of the input image. To input the features into the transformer encoder, we first split the input image into 9 patches, $x_i \in \mathbb{R}^{16\times 16\times 1}$, $i = $

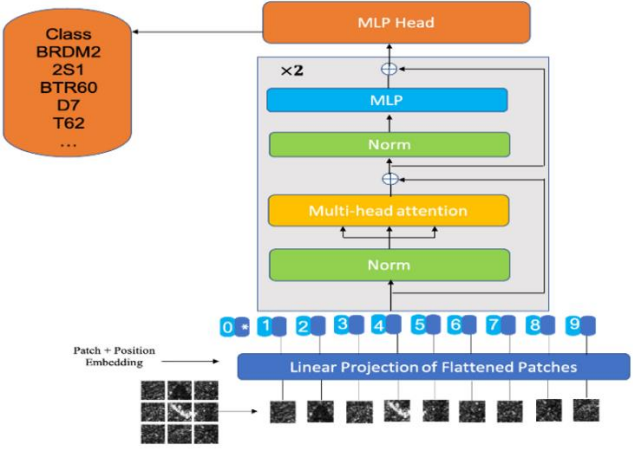

Fig. 2. Overview of the proposed architecture.

{1, …, N}, where N $= \frac{48\times 48}{16^2} = 9$ is the number of patches and 16 is the patch size. Then, the 9 patches are flattened into a sequence of 1D patches, $x_i \in \mathbb{R}^{16^2\times 1}$. Next, the flattened patches are mapped to 512 dimensions with a trainable linear projection (Eq.1). The output of this linear projection $Z_i \in \mathbb{R}^{D\times 1}$, $i = \{1, …, 9\}$, refers to the patch embeddings. Besides, we pretend a learnable embedding($Z_0$=Xclass) to the patch embeddings.

To utilize the position information of each patch of the image, we add the position encodings $P_i \in \mathbb{R}^{D\times 1}$, $i = \{0,1, …, 9\}$, for each patch $Z_i$. Then, $P_i + Z_i$ are inputted into transformer encoder directly (Eq.2). The transformer encoder consists of two identical encoder layers. Each encoder layer has a multi-head self-attention and feed-forward network (Eq.3,4). The output of the encoder $f_0 \in \mathbb{R}^{D\times 1}$ acts as input of MLP for image classification (Eq. 5). The calculation can be formulated as:

$$Z_i = W \cdot x_i, W \in \mathbb{R}^{K*16^2}, i = \{1, …, 9\} \quad (1)$$

$$y_0 = [P_0 + Z_0, P_1 + Z_1, …, P_9 + Z_9] \quad (2)$$

$$y'_l = MSA(LN(y_{l-1})) + y_{l-1}, \quad l = 1,…,L \quad (3)$$

$$y_l = MLP(LN(y'_l)) + y'_l, \quad l = 1,…,L \quad (4)$$

$$y_L = [f_0, f_1, f_2, …, f_9] \quad (5)$$

Where l denotes the number of encoder layers in the encoder, MSA denotes the multi-head self-attention mechanism, and FFN denotes the feed-forward network.

### C. Fine-tuning

This paper selected multi-categorical cross-entropy as a loss function which is optimized by a self-decaying Adam optimizer with a learning rate starting from 0.001. We trained the model for 80 epochs with a batch size of 64. The LViT architecture is fine-tuned to 2 layers and 2 heads, an embedding size of 256, and a drop-out rate of 0.3 was applied, it is noticeable that this ensemble of gradient steps, embedding size, and heads are found after many trials which is helpful in aggregating the extracted features to achieve better SAR recognition results. Besides, we started the training process with our generally optimized initial parameters (GOIP) which is derived by fine-tuning the LViT with existing weights on SAR-relevant tasks (such as the optical images of SAR targets and other general SAR imageries). This training strategy has been proven to be powerful in other similar downstream experiments [13] [14] [15], and more style-transferred details about this training methodology can be found in [16].

In summary, Section 2 provides a comprehensive overview of the dataset used, the architecture of the model, and the fine-tuning approach. It begins with a detailed description of the MSTAR Data Set, which includes ten categories of targets such as bulldozers, tanks, armored carriers, howitzers, anti-air units and trucks. The data, gathered by a plane using SAR imagery technology, presents challenges due to similarities among target classes and variations in depression angles between training and testing sets, emphasizing the need for a model with strong generalization abilities and the concept of meta-learning with expandable K-way and n-shots. This section sets a solid foundation for the results and analysis presented in Section 3, where the effectiveness of the LViT model in classifying SAR images is demonstrated and compared with traditional methods.

### III. RESULTS AND ANALYSIS

#### A. Classification in Progress

The entire experiment was supported via a laptop with the Intel(R) i7-11800H CPU, and a 16 GB NVIDIA RTX 3080 Max-Q Laptop GPU. Heatmaps in Fig.3 that have genuine target labels in the row and predicted target labels in the column demonstrate the recognition process and classification performance of the constructed LViT.

From the 4 typical stages of the classification process, it can be observed that all images were gradually classified, starting from an initial random stage (where most test results are set randomly), the stages show that the predicted outcomes gradually converging, ultimately leading to a well-trained state where the most of predictions fall along the diagonal. This demonstrates that the model performs effectively in identifying these different categories.

When looking inside the confusion matrix, it is interesting to find that some quasi-rectangles are formed in both initial and intermediate stages (the top left and bottom right of the heatmaps), these blocks generally indicate the similarity among SAR images of different categories. For example, the first block indicates that two types of armored carriers confuse with T-62 tank and the howitzer, and different sub-versions of T-72 tanks (SN 132, SN C71) and the BMP2 armored carrier confuse each other. It is noticeable that since we applied the GOIP and a training method introduced in Section II-C, some quasi blocks can even be found in the beginning stage, these blocks will typically occur in the intermediate training epochs if we did not include the GOIP.

The final heatmap shows that the most amount of predicted labels correctly match the ground truth (in blue or dark blue). Except for the ZSU 23/4 and ZIL 131 rows, every remaining box is in the light color which means only few predicted labels are wrongly categorized into other categories. Fig.4. shows the final classification heatmap of our network, from where the detailed quantity of each class can be clearly found and a further result analysis based on this heatmap will be conducted in the following section.

### B. Results in Confusion Matrix

A more detailed analysis of each category is shown in Fig. 4. below, from which it can be observed that, for the LViT network, an overall recognition rate of 97.75% is achieved. And except for the ZSU23/4, and ZIL-131 classes, all the remaining 8-class targets are in good recognition with each accuracy exceeding 95% (this network even achieved full recognition for the D-7 and BRDM2).

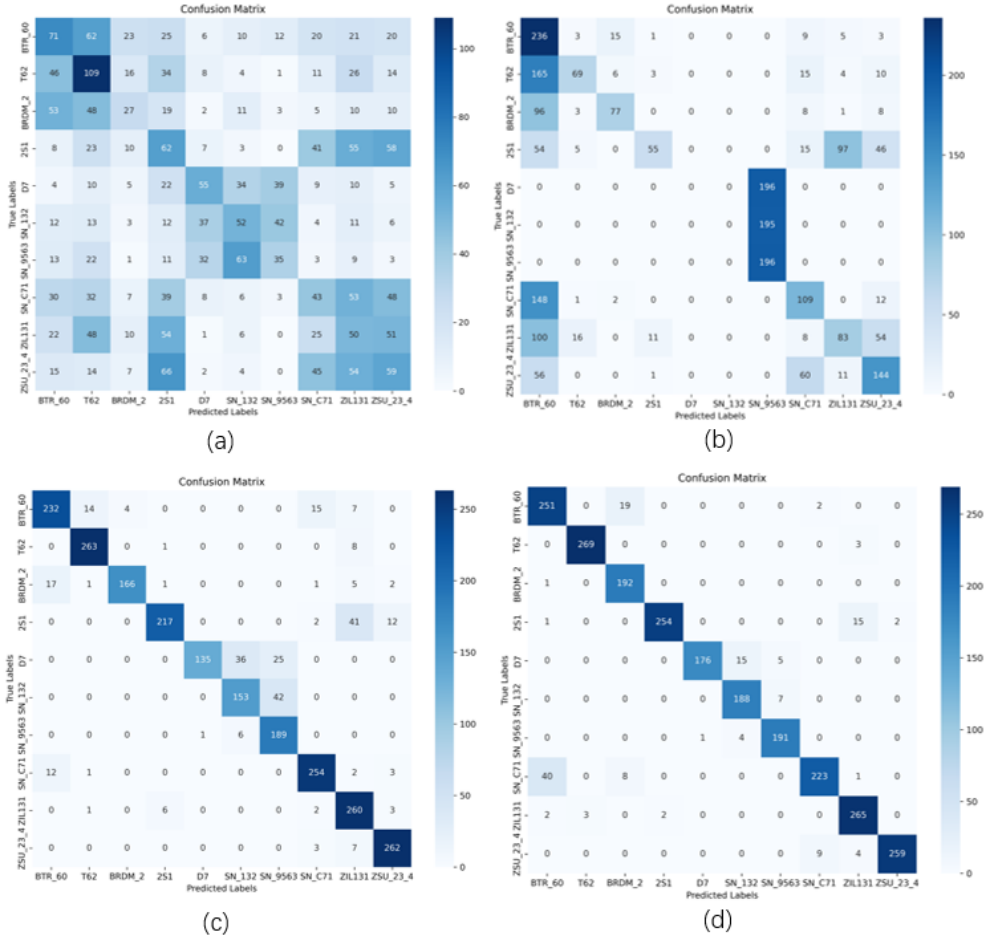

Fig. 3. The feature extraction and learning process in different model training epochs (a) Beginning phase (when epoch=0). (b) Incipient classification results (when epoch=30). (c) Advanced feature extraction results (when epoch=50). (d) Approaching ultimate well-trained stage (when epoch=80).

When looking inside each category, it is observed that for the ZIL-131: 7.35% of 2S1 were thought to be BTR70 (sn-c71) and 1.1% of T62 were misunderstood as 2S1 by the LViT network, it is also noticed that some part of T72 were recognized as D7 bulldozer, this might due to all these three types of targets have 'rectangular appearance' which makes the network hard to distinguish among them. As for the worst recognized ZSU23/4 anti-aircraft gun, 5.32% of this target was wrongly recognized as a BTR70, this might be because they both have similar 'round turret', especially when the plane imagines the target from the top. Therefore, the network gets very confused.

For the three types of armored carriers (BTR70, BTR60, BMP2), although they all belong to the carrier category, the network is not perplexed by shared similarities, on the contrary, few of them are wrongly recognized with others, and they are all classified with an over 95.97% accuracy. It was also worth pointing out that the BMP2 could be confused with tanks because of similar gun barrels. The other categories like the T72 tank and BRDM2 amphibious armoured scout also have very appreciating results when adopting the LViT network.

### C. Model Evaluation

Recall, Precision, and F1-score [14] are traditional indicators for evaluating the model. In terms of our 10-class classification task, the overall recall value is the average of all the 10 classes and refers to the proportion that true positive classified samples take within the pool of all expected results, and the overall precision is the average proportion that true positive results take within the domain of positive results. Similarly, the overall F1-Score represents the average result of the so-called 'harmonic aver-age' of the 10-class targets' recall and precision, which in general has a positive correlation with the classification ability of a model. The mathematical representations for all these three parameters are [7] [8]:

$$Precision = \frac{\sum_{c=1}^{N} precision_c}{N}, precision_c = \frac{TP_c}{TP_c+FP_c} \quad (6)$$

$$Recall = \frac{\sum_{c=1}^{N} recall_c}{N}, recall_c = \frac{TP_c}{TP_c+FN_c} \quad (7)$$

$$F1\_score = \frac{\sum_{c=1}^{N} f1_c}{N}, f1_c = \frac{2 \times precision_c \times recall_c}{precision_c+recall_c} \quad (8)$$

Where Capital Precision, Recall, and F1 score represent the corresponding overall results of the 10 classes, and $precision_c$, $recall_c$, $f1_c$ represent the separate results associated with each category ranging from 0 to 9. TP is the number of true positives (both real and predicted labels are positive), FP is the number of false positives (the real label is negative, but the predicted label is positive), TN is the number of true negatives (the real true label is predicted as negative), and FN is the number of the false negatives (the real label is positive, but the predicted label is negative). It should be stressed that these four parameters rely on the pre-designed threshold values which categorize the negative and positive values by defining whether labels with a certain confidence score be classified as positive or negative. And in Python, we can use the 'sklearn' to calculate these evaluation parameters. The calculated results of our investigated model show that this model has a recall of 97.42%, a precision of 97.49%, and an F1-score of 97.45%.

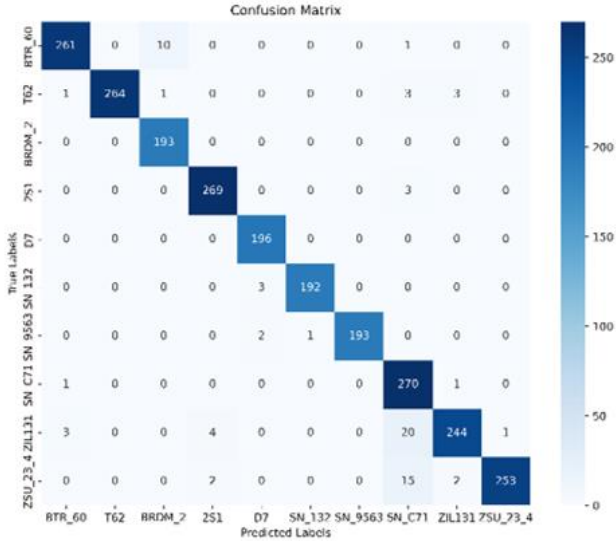

Fig. 4. The final classification results of the model in heatmap.

*D. Results Comparison*

Fig. 5. selected some typical experiments throughout these years including conventional structure-based methods like CDSPP [17] and deep learning-based methods such as CNN-SVM [18], Multi-kernel/Multiview-based structure [8], autoencoder and its variant [2]. It can be found that in comparison with these referred structures, the LViT structure achieved enhanced recognition rates for 4 categories out of 10 together with a better overall performance. Considering the fact that all outcomes were derived from a structure with lighter layers and without any convolution layers, this result is satisfying [19] [20]. When looking inside each category, it is also found that except for the ZSU23/4 and ZIL 131 category, the remaining classes generally achieved good results especially the 2 kinds of armored targets (BTR70, BRDM2) as well as the 2S1 howitzer and the truck (over 98%) [21] [22]. It should also be pointed out that the LViT is expandable in layers and can be further applied to deal with other large scene downstream problems or the same SAR image classification task but with more input variants [23] [24]. Its promising power in dealing with the future huge data sets has already been demonstrated in [25] [26].

| Methods | CDSPP [17] | CNN-SVM [18] | Autoencoder [2] | EDR-Autoencoder[2] | MKSFF-CNN [8] | LViT |
|---|---|---|---|---|---|---|
| ZSU23/4 | 97.81 | 96.35 | 87.34 | 94.53 | **97.81** | 89.73 |
| 2S1 | 88.69 | 82.12 | 90.39 | 93.80 | 93.80 | **98.89** |
| BMP2 | 96.94 | 96.94 | 88.96 | 92.86 | 94.36 | **98.47** |
| BTR70 | 80.58 | 78.46 | 90.05 | 87.90 | **99.49** | 99.27 |
| T72 | 94.87 | 89.74 | 89.10 | 91.79 | **100.00** | 98.47 |
| BTR60 | 85.74 | **100.00** | 69.85 | 79.55 | 98.46 | 95.97 |
| D7 | 97.08 | 95.99 | 96.08 | 98.91 | 99.27 | **100.00** |
| T62 | 98.18 | 97.08 | 77.15 | **99.64** | 95.24 | 97.06 |
| BRDM2 | 95.99 | 89.78 | 92.12 | 96.72 | 97.45 | **100.00** |
| ZIL131 | 94.87 | 90.11 | 95.15 | 94.14 | **99.27** | 89.71 |
| Average | 91.01 | 91.66 | 87.62 | 91.29 | 97.44 | **97.75** |

Fig. 5. The Comparison of Different Methods.

IV. CONCLUSION AND FUTURE WORK

In conclusion, this study has successfully demonstrated the effectiveness of deep learning-based methods, particularly the lightweight vision transformer (LViT), in enhancing Synthetic Aperture Radar (SAR) Automatic Target Recognition (ATR). This approach has shown significant advantages over traditional network structures like CNNs and autoencoders in terms of recognition accuracy and robustness. Deep learning-based methods can benefit the SAR ATR in terms of making researchers free from designing sophisticated feature extraction algorithms. Over the years, many effective neural network structures have been proposed and make the SAR ATR field move forward to the peak. However, most structures rely on networks or algorithms focusing on exploring local or sequence patterns of SAR images while partially overlooking potential global patterns in the SAR ATR task. In this paper, we applied and tested the power of LViT in classifying the MSTAR data set which shows that a global pattern-focused methodology can achieve both good recognition results and robust behavior. For future work, we plan to further advance our research by incorporating multi-view data collection methods, which are expected to enrich the dataset with more diverse and comprehensive perspectives, and with style-transfer inputs such as thermal, optical, and segmented representations of the same target, thereby improving the model's ability to generalize across different scenarios. Additionally, we aim to integrate deep learning uncertainty metrics into our model and test the structure in more downstream tasks. This integration will provide a more nuanced understanding of the model's confidence in its predictions, potentially leading to more reliable and interpretable results in SAR image classification. These future endeavors will not only refine our current achievements but also pave the way for more sophisticated and efficient SAR ATR mechanisms.